\documentclass{article}

\usepackage{microtype}
\usepackage{graphicx}
\usepackage{subfigure}
\usepackage{booktabs} % for professional tables

\usepackage{hyperref}

\usepackage[accepted]{icml2024}
\usepackage{amsmath}
\usepackage{amssymb}
\usepackage{mathtools}
\usepackage{amsthm}
\usepackage{tabularx}
\usepackage{rotating}
\usepackage{adjustbox}
\usepackage{vcell}
\usepackage{longtable}
\newcolumntype{L}[1]{>{\raggedright\arraybackslash}p{#1}}
\newcolumntype{C}[1]{>{\centering\arraybackslash}p{#1}}
\usepackage[capitalize,noabbrev]{cleveref}
\usepackage{comment}
\theoremstyle{plain}

\theoremstyle{definition}

\theoremstyle{remark}

\usepackage[textsize=tiny]{todonotes}

\icmltitlerunning{ICML 2024 Topological Deep Learning Challenge}

\begin{document}

\twocolumn[
\icmltitle{ICML Topological Deep Learning Challenge 2024: Beyond the Graph Domain}



\icmlsetsymbol{equal}{*}

\begin{icmlauthorlist}
\icmlauthor{Guillermo Bern\'ardez}{equal,ChO,ChR}
\icmlauthor{Lev Telyatnikov}{equal,ChO,ChR}
\icmlauthor{Marco Montagna}{ChO,ChR}
\icmlauthor{Federica Baccini}{ChO,ChR}
\icmlauthor{Mathilde Papillon}{ChC,ChR}
\icmlauthor{Miquel Ferriol-Galm\'es}{ChC}
\icmlauthor{Mustafa Hajij}{ChC,ChR}
\icmlauthor{Theodore Papamarkou}{ChC,ChR}
\icmlauthor{Maria Sofia Bucarelli}{ChR}
\icmlauthor{Olga Zaghen}{ChR}
\icmlauthor{Johan Mathe}{ChR}
\icmlauthor{Audun Myers}{ChR}
\icmlauthor{Scott Mahan}{ChR}
\icmlauthor{Hansen Lillemark}{ChR}
\icmlauthor{Sharvaree Vadgama}{ChC}
\icmlauthor{Erik Bekkers}{ChC}
\icmlauthor{Tim Doster}{ChC}
\icmlauthor{Tegan Emerson}{ChC}
\icmlauthor{Henry Kvinge}{ChC}
% Participants: 
\icmlauthor{Katrina Agate}{challenge_participant}
\icmlauthor{Nesreen K Ahmed}{challenge_participant}
\icmlauthor{Pengfei Bai}{challenge_participant}
\icmlauthor{Michael Banf}{challenge_participant}
\icmlauthor{Claudio Battiloro}{challenge_participant}
\icmlauthor{Maxim Beketov}{challenge_participant}
\icmlauthor{Paul Bogdan}{challenge_participant}
\icmlauthor{Martin Carrasco}{challenge_participant}
\icmlauthor{Andrea Cavallo}{challenge_participant}
\icmlauthor{Yun Young Choi}{challenge_participant}
\icmlauthor{George Dasoulas}{challenge_participant}
\icmlauthor{Matouš Elphick}{challenge_participant}
\icmlauthor{Giordan Escalona}{challenge_participant}
\icmlauthor{Dominik Filipiak}{challenge_participant}
\icmlauthor{Halley Fritze}{challenge_participant}
\icmlauthor{Thomas Gebhart}{challenge_participant}
\icmlauthor{Manel Gil-Sorribes}{challenge_participant}
\icmlauthor{Salvish Goomanee}{challenge_participant}
\icmlauthor{Victor Guallar}{challenge_participant}
\icmlauthor{Liliya Imasheva}{challenge_participant}
\icmlauthor{Andrei Irimia}{challenge_participant}
\icmlauthor{Hongwei Jin}{challenge_participant}
\icmlauthor{Graham Johnson}{challenge_participant}
\icmlauthor{Nikos Kanakaris}{challenge_participant}
\icmlauthor{Boshko Koloski}{challenge_participant}
\icmlauthor{Veljko Kovač}{challenge_participant}
\icmlauthor{Manuel Lecha}{challenge_participant}
\icmlauthor{Minho Lee}{challenge_participant}
\icmlauthor{Pierrick Leroy}{challenge_participant}
\icmlauthor{Theodore Long}{challenge_participant}
\icmlauthor{German Magai}{challenge_participant}
\icmlauthor{Alvaro Martinez}{challenge_participant}
\icmlauthor{Marissa Masden}{challenge_participant}
\icmlauthor{Sebastian Mežnar}{challenge_participant}
\icmlauthor{Bertran Miquel-Oliver}{challenge_participant}
\icmlauthor{Alexis Molina}{challenge_participant}
\icmlauthor{Alexander Nikitin}{challenge_participant}
\icmlauthor{Marco Nurisso}{challenge_participant}
\icmlauthor{Matt Piekenbrock}{challenge_participant}
\icmlauthor{Yu Qin}{challenge_participant}
\icmlauthor{Patryk Rygiel}{challenge_participant}
\icmlauthor{Alessandro Salatiello}{challenge_participant}
\icmlauthor{Max Schattauer}{challenge_participant}
\icmlauthor{Pavel Snopov}{challenge_participant}
\icmlauthor{Julian Suk}{challenge_participant}
\icmlauthor{Valentina S\'anchez}{challenge_participant}
\icmlauthor{Mauricio Tec}{challenge_participant}
\icmlauthor{Francesco Vaccarino}{challenge_participant}
\icmlauthor{Jonas Verhellen}{challenge_participant}
\icmlauthor{Frederic Wantiez}{challenge_participant}
\icmlauthor{Alexander Weers}{challenge_participant}
\icmlauthor{Patrik Zajec}{challenge_participant}
\icmlauthor{Blaž Škrlj}{challenge_participant}
\icmlauthor{Nina Miolane}{ChO}
\end{icmlauthorlist}

\icmlaffiliation{ChO}{Challenge Organizer}
\icmlaffiliation{ChR}{Challenge Reviewer}
\icmlaffiliation{ChC}{Challenge Contributor}
\icmlaffiliation{challenge_participant}{Challenge Participant}

% \icmlaffiliation{diag}{Department of Computer, Control, and Management Engineering ``Antonio Ruberti", Università degli Studi di Roma ``La Sapienza", Via Ariosto, 25, \\Roma, 00185, Italy}

\icmlcorrespondingauthor{Guillermo Bernárdez}{guillermo\_bernardez@ucsb.edu} 

\icmlcorrespondingauthor{Lev Telyatnikov}{lev.telyatnikov@uniroma1.it}

\icmlkeywords{Topological Deep Learning, Challenge, Machine Learning, ICML}

\vskip 0.3in
\editorsListText
\vskip 0.3in
]



\printAffiliationsAndNotice{\icmlEqualContribution} % otherwise use the standard text.

\begin{abstract}
This paper describes the 2nd edition of the \textit{ICML Topological Deep Learning Challenge} that was hosted within the ICML 2024 ELLIS Workshop on Geometry-grounded Representation Learning and Generative Modeling (GRaM). The challenge focused on the problem of representing data in different discrete topological domains in order to bridge the gap between Topological Deep Learning (TDL) and other types of structured datasets (e.g. point clouds, graphs). Specifically, participants were asked to design and implement topological liftings, i.e. mappings between different data structures and topological domains --like hypergraphs, or simplicial/cell/combinatorial complexes. The challenge received 52 submissions satisfying all the requirements. This paper introduces the main scope of the challenge, and summarizes the main results and findings. 
\end{abstract}

\section{Introduction}
\label{intro}

\begin{figure*}[!ht]
    \centering
    \includegraphics[width=0.9\textwidth]{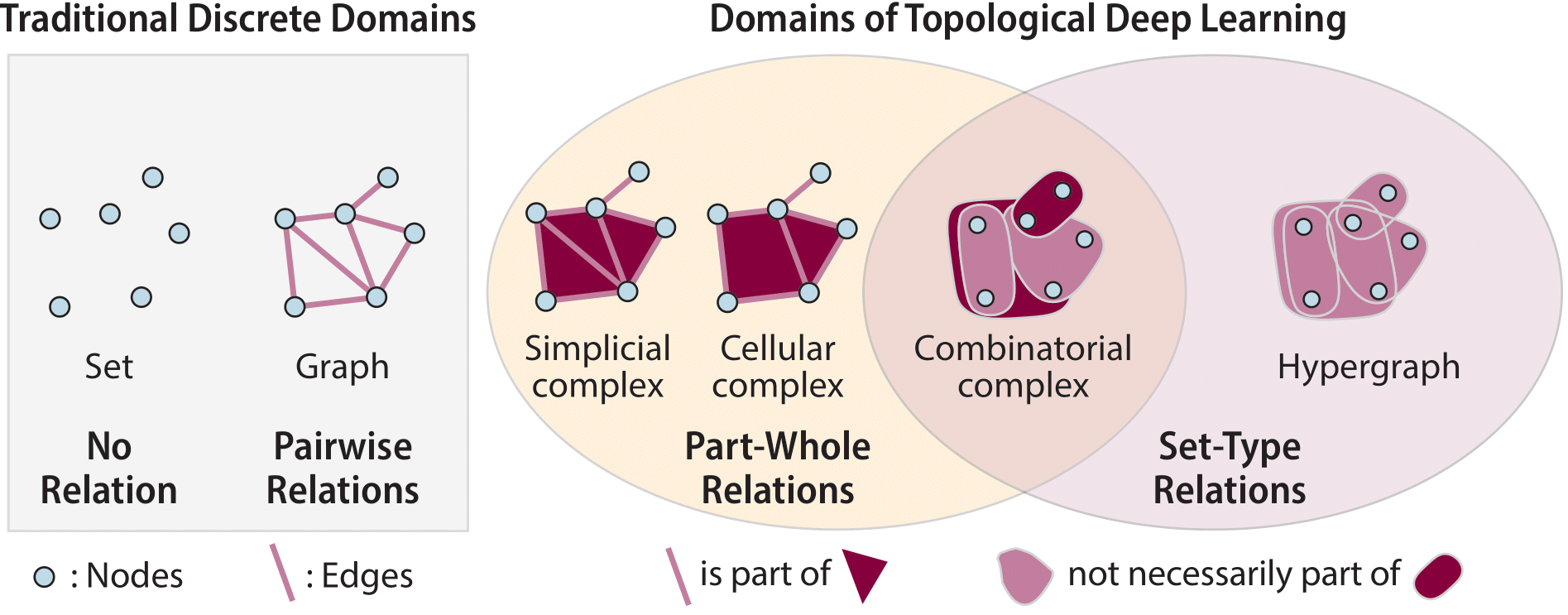}
    \caption{Different discrete domains. Figure adopted from \cite{papillon2023architectures}.}
    \label{fig:domains}
\end{figure*}

The field of Topological Deep Learning (TDL) aims to extend Graph Neural Networks (GNN) \cite{scarselli2009} by naturally processing relations between two or more elements~\cite{hajij2022topologicalDL, papillon2023architectures, hajij2021topological,bodnar2021weisfeiler,bodnar2021weisfeilercellular}. %, potentially characterizing interactions in realistic complex systems \cite{hajij2022topologicalDL, papillon2023architectures, hajij2021topological,bodnar2021weisfeiler,bodnar2021weisfeilercellular}. 
In particular, TDL methods allow to go beyond the paradigm of pairwise interactions by encoding higher-order relationships using algebraic topology concepts \cite{bick2023higher,bodnar2023topological,battiloro2023topological}. Fig. \ref{fig:domains} presents a visual comparison of traditional discrete domains (i.e. pointclouds, graphs) versus the standard discrete topological domains used to model $n$-body relations (simplicial/cellular/combinatorial complexes, hypergraphs). %The introduction of suitable methods to process higher-order structures allows to study $n$-body interactions existing in most complex systems, thus enriching the information obtained by only looking at the pairwise relations \cite{baccini2022weighted}. 

Despite its recent emergence, TDL is already postulated to become a relevant tool in many research areas and applications, from complex physical systems \cite{battiston2021physics} and signal processing \cite{barbarossa2020topological} to molecular analysis \cite{bodnar2021simplicial} and social interactions \cite{schaub2020random}, to name a few. However, a current limiting factor in the extensive use of higher-order structures is that most datasets are usually stored as pointclouds or graphs. %As a consequence, it is unclear how a higher-order structure that encodes the information present in such data can be extracted \cite{telyatnikov2024topobenchmarkx}. 
Although researchers have introduced various mechanisms for extracting higher-order elements (e.g. \citet{xu2022groupnet,battiloro2023latent,bernardez2023topological,elshakhs2024comprehensive,hajij2022higher,hoppe2024representing}), it remains unclear how to optimize the process given a specific dataset and task.

\begin{figure*}[!ht]
    \centering
    \includegraphics[width=0.9\textwidth]{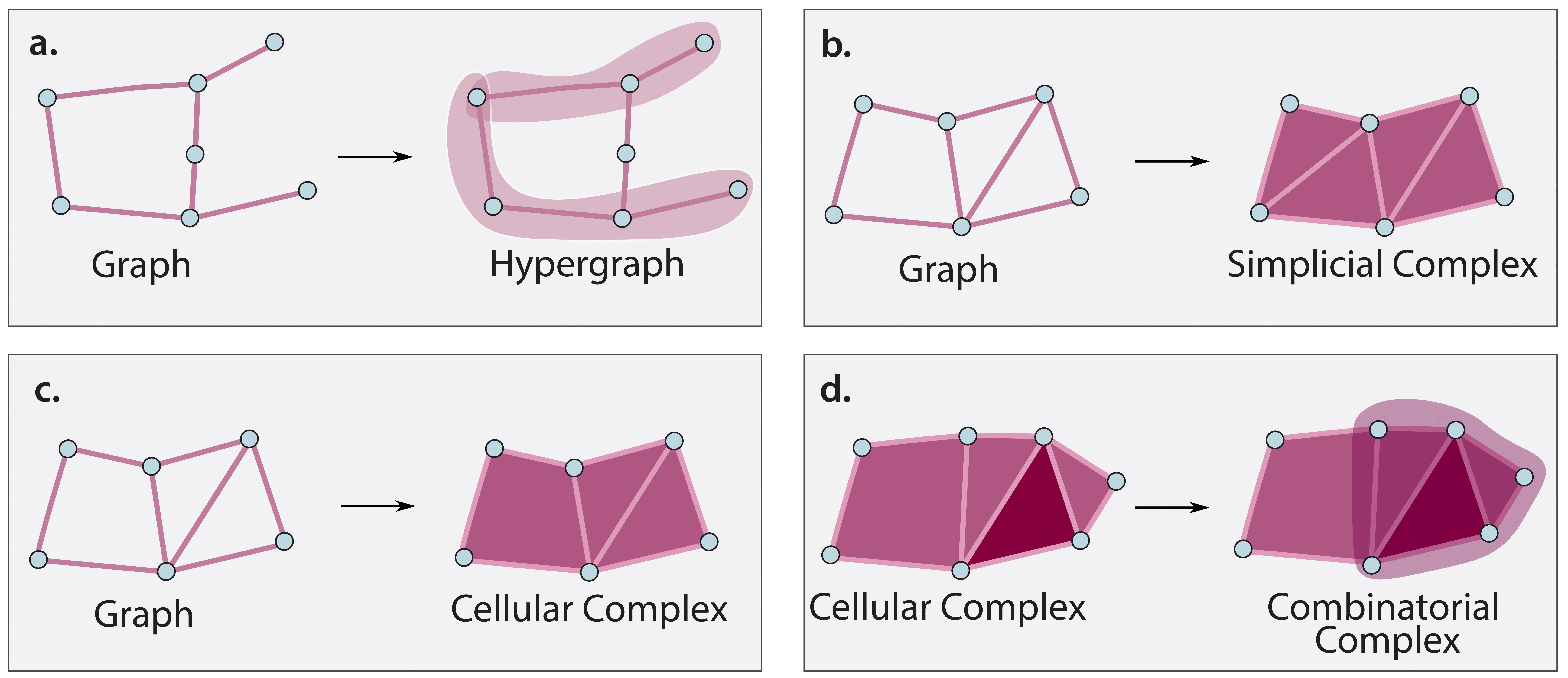}
    \caption{Examples of liftings: (a) A graph is lifted to a hypergraph by adding hyperedges that connect groups of nodes. (b) A graph can be lifted to a cellular complex by adding faces of any shape. (c) Hyperedges can be added to a cellular complex to lift the structure to a combinatorial complex. Figure adopted from \cite{hajij2023combinatorial}}
    \label{fig:lifting}
\end{figure*}

In this context, developing tools and methods to move between different discrete (topological) domains represents one of the most pressing open challenges in TDL~\cite{papamarkou2024position}. The process of mapping a data structure to a different topological domain is formalized using the concept of \textit{topological lifting} (or, equivalently, \textit{lifting}) \cite{papillon2023architectures}. %Several liftings have been proposed in the literature, each one involving different types of topological structures \cite{battiloro2023latent,elshakhs2024comprehensive,hajij2022higher,hoppe2024representing,jogl2022reducing}. 
Additionally, a \textit{feature lifting} is a particular lifting that transfers data from an original domain where a signal (node/edge features) exists, to a new domain where new topological structures can emerge, such as simplicial/cell complexes. Fig. \ref{fig:lifting} depicts some visual examples of different liftings involving several topological domains. 

In this framework, the main purpose of the \textit{ICML 2024 Topological Deep Learning Challenge} is to foster new research and knowledge about effective liftings between different topological domains and data structures. The introduction of new topological liftings between different topological domains, as well as an efficient implementation of both new and existing mappings, might help to enforce the impact of TDL to a broader range of use cases and scenarios.

Besides reporting the main organizational aspects of the challenge, this paper aims to give an overview of the main results achieved by participants. 

%\paragraph{Remark:} Consistent with the aims of an open environment for sharing participation in this activity is completely voluntary and no support or endorsement of any of the participating parties by any of the other participating parties is provided. All submissions are the views of the individual participants only and should be taken, as is with all faults and without any guarantee, promise or endorsement of any kind.

\section{Setup of the Challenge}
% \textcolor{blue}{where is the challenge held, Scope of the challenge, available resources (generic. Insert link to webpage od the challenge), }
The challenge\footnote{Website: \url{https://pyt-team.github.io/packs/challenge.html}} was hosted by the Geometry-grounded Representation Learning and Generative Modeling (GRaM) Workshop\footnote{Workshop website: \url{https://gram-workshop.github.io/}} at the International Conference on Machine Learning (ICML) 2024.
Participants were asked to implement a topological lifting, apply it to a toy dataset, and test the results with an existing provided model for the considered target domain.

\subsection{Guidelines}
Participation was free and open to everyone --only principal PyT-Team developers were excluded. %\footnote{However, US researchers are not allowed by law to co-author papers with scholars from some countries and institutions. Unfortunately, in these cases participants wouldn't be eligible for the publication-based awards, even in the case of winning the competition.} 
To enroll in the challenge it was sufficient to:
\begin{itemize}
    \item Send a valid Pull Request (PR) --i.e. passing all tests-- before the deadline.
    \item Respect Submission Requirements (see below).
\end{itemize}
Teams were accepted, and there was no restriction on the number of team members. An acceptable PR automatically subscribed a participant/team to the challenge. A Pull Request could not contain more than one lifting. However, there was no restriction on the number of submissions (i.e. PRs) per participant/team. 

Consistent with the aims of an open environment for sharing participation, in this activity was completely voluntary and no support or endorsement of any of the participating parties by any of the other participating parties was provided. All submissions are the views of the individual participants only and should be taken, as is with all faults and without any guarantee, promise or endorsement of any kind.

\subsection{Submission Requirements}\label{subs::subreq}
The submission had to have a valid lifting transformation between any pair of the following data structures: point-cloud/graph, hypergraph, simplicial complex, cell complex, and combinatorial complex. For a lifting to be valid, participants had to implement a mapping between the topological structures of two of the considered domains --\textit{topological lifting}. Participants may optionally implement a procedure to define the features over the resulting topology --\textit{feature lifting}.

\subsubsection*{Topology Lifting (Required)}

Submissions could implement already proposed liftings from the literature, as well as propose novel approaches. In the case of original liftings, we note that neither the challenge nor its related publications would prevent participants from publishing their work: they keep all the credit for their implementations.

%Despite liftings from point-cloud/graph data to higher-order topological domains are encouraged, we remark that any combination is valid.

For a lifting from a certain source domain \texttt{src} (e.g. graph) to a topological destination \texttt{dst} (e.g. simplicial), a submission consisted of a PR to the ICML Challenge repository with the following files:

\begin{enumerate}
    \item A Python script implementing the topological lifting in a single class using the provided \texttt{\{src\}2\{dst\}} transform primitives.

    \item A configuration file defining the default parameters of the implemented transform.

    \item A Jupyter notebook that loads a dataset from the \texttt{src} domain, applies the implemented lifting to transform the data into the \texttt{dst} domain, and runs a model from \texttt{TopoModelX}~\cite{} over the lifted dataset.

    \item A Python script which contains the unit tests for all implemented methods and classes.
\end{enumerate}

\subsubsection*{Feature Lifting (Optional)}

Some TDL models require well-defined features on higher-order structures (e.g. 2-cells, hyperedges); therefore, in their more general formulation, liftings also need to produce initial features for every topological element of the target domain. %In particular, in all our examples we make use of a straightforward \texttt{SumProjection} transform to that end, which gets the desired structural features by sequentially projecting the original signals via incidence matrices.
Participants were more than welcome to implement new feature liftings mappings, %which could be added to the \texttt{feature\_liftings.py} file at the \texttt{modules/transforms/feature\_liftings/} directory. However, 
although it was optional and only regarded as a bonus.

\subsection{Award Categories}

Given the lack of an exhaustive analysis of different types of procedures to infer the topological structure within TDL, there was no particular requirement for submitted liftings --apart from a high-quality code implementation. To promote and guide diversity in submissions, we proposed 4 general, non-mutually exclusive Award Categories (ACs) according to the following 2 taxonomies.

\paragraph{By target domain:}
\begin{itemize}
    \item \textbf{1st AC:} Best implementation of a lifting to Simplicial of Cell Domain.
    \item \textbf{2nd AC:} Best implementation of a lifting to Combinatorial Complex, Hypergraph, or Graph Domain.
\end{itemize}

\paragraph{By leveraged information:}
\begin{itemize}
    \item \textbf{3rd AC:} Best feature-based lifting, including liftings that leverage both the graph connectivity simultaneously. 
    \item \textbf{4th AC:} Best implementation of a lifting using exclusively the graph connectivity.
\end{itemize}

In addition to the winners' award categories, the challenge also featured honorable mentions. These honorable mentions were determined based on aggregated reviewers' comments.

%some aspects like originality, theoretical robustness, loading interesting datasets, implementing new feature liftings, etc

\subsection{Evaluation Method}
The Condorcet method was used to rank the submissions and decide on the winners in each category. The evaluation criteria were:
\begin{itemize}
    \item Does the submission implement the lifting correctly? Is it reasonable and well-defined?
    \item How readable/clean is the implementation? How well does the submission respect the submission requirements?
    \item Is the submission well-written? Do the docstrings clearly explain the methods? Are the unit tests robust?
\end{itemize}
Note that these criteria did not reward the final model performance nor the complexity of the method. Rather, the goal was to implement well-written and accurate liftings that would unlock further experimental evidence and insights in this field.

Selected PyT-Team maintainers and collaborators, as well as each team whose submission(s) respect(s) the guidelines, voted once to express their preference for the best submission in each category. Note that each team was allowed only one vote per category, regardless of the number of team members. %Note that each team only got one vote/domain, even if there were several participants in the team.

\subsection{Software Practices}
All submitted code had to comply with the challenge's GitHub Action workflow, successfully passing all tests, linting, and formatting (i.e., ruff). Moreover, to ensure consistency, we asked participants to use TopoNetX's classes to manage simplicial/cell/combinatorial complexes whenever these topological domains are the target --i.e., destination-- of the lifting.
Moreover, we highly encouraged the use of TopoX~\cite{hajij2024topox} and NetworkX~\cite{hagberg2020networkx} libraries when possible.

\section{Submissions and Winners}

The challenge received 56 submissions in total, out of which 52 were valid according to the requirements (see Subsection \ref{subs::subreq}). Regarding these valid submissions, they come from 31 different teams, with together sum up a total of 57 participants. Each award category accounted for 24, 28, 25, and 27 submissions correspondingly. Table \ref{tab:submissions} lists all qualifying submissions. 

The winners were announced publicly at the ICML Workshop on Geometry-grounded Representation Learning and Generative Modeling, on social media, as well as on the official challenge website. Regardless of the final rankings, we want to emphasize that all the submissions were of exceptionally high quality. We warmly congratulate all participants.

\subsection{Award Category Winners}

Three prizes were awarded for each category, corresponding to first, second and third places in the Condorcet voting ballot results.

\paragraph{1st AC:} Best lifting to Simplicial of Cell Domain.
\begin{enumerate}
    \item \textit{Random Latent Clique Lifting} (Graph to Simplicial) by Mauricio Tec, Claudio Battiloro, George Dasoulas. 
    \item \textit{Hypergraph Heat Kernel Lifting} (Hypergraph to Simplicial) by Matt Piekenbrock. 
    \item \textit{DnD Lifting} (Graph to Simplicial) by Jonas Verhellen
\end{enumerate}

\paragraph{2nd AC:} Best lifting to Graph, Hypergraph, or Combinatorial Domain.
\begin{enumerate}
    \item \textit{Simplicial Paths Lifting} (Graph to Combinatorial) by Manuel Lecha, Andrea Cavallo, Claudio Battiloro. 
    \item \textit{Matroid Lifting} (Graph to Combinatorial) by Giordan Escalona. 
    \item \textit{Forman-Ricci Curvature Coarse Geometry Lifting} (Graph to Hypergraph) by Michael Banf, Dominik Filipiak, Max Schattauer, Liliya Imasheva. 
\end{enumerate}

\paragraph{3rd AC:} Best feature-based lifting.
\begin{enumerate}
    \item \textit{PointNet++ Lifting} (Pointcloud to Hypergraph) by Julian Suk, Patryk Rygiel. 
    \item \textit{Kernel Lifting} (Graph to Hypergraph) by Alexander Nikitin. \item \textit{Mixture of Gaussians + MST Lifting} (Pointcloud to Hypergraph) by Sebastian Mežnar, Boshko Koloski, Blaž Škrlj.
\end{enumerate}

\paragraph{4th AC:} Best connectivity-based lifting.
\begin{enumerate}
    \item \textit{Matroid Lifting} (Graph to Combinatorial) by Giordan Escalona. 
    \item \textit{Forman-Ricci Curvature Coarse Geometry Lifting} (Graph to Hypergraph) by Michael Banf, Dominik Filipiak, Max Schattauer, Liliya Imasheva. 
    \item \textit{Hypergraph Heat Kernel Lifting} (Hypergraph to Simplicial) by Matt Piekenbrock.
\end{enumerate}

\subsection{Honorable Mentions}

Apart from voting for the best implementations in each award category, reviewers were also asked to highlight submissions and/or participants if they found their implementations particularly interesting. Given the high quality of received submissions, reviewers's feedback originated two extra honorable mentions' categories:

\paragraph{Great Contributors:} Teams or participants that have submitted several top quality liftings, becoming great contributors of this project.

\begin{itemize}
    \item Martin Carrasco (PRs 28, 29, 41, 50). 
    \item Bertran Miquel-Oliver, Manel Gil-Sorribes, Alexis Molina, Victor Guallar (PRs 14, 16, 21, 37, 42). 
    \item Theodore Long (PRs 22, 35, 65).
    \item Jonas Verhellen (PRs 5, 7, 8, 10, 11).
    \item Pavel Snopov (PRs 6, 9, 18, 20).
    \item Julian Suk, Patryk Rygiel (PRs 23, 34, 53).
\end{itemize}

\paragraph{Highlighted Submissions:} Original and/or outstanding submissions.

\begin{itemize}
    \item \textit{Modularity Maximization Lifting} (Graph to Hypergraph) by
Valentina Sánchez.
    \item \textit{Universal Strict Lifting} (Hypergraph to Combinatorial) by
Álvaro Martínez.
    \item \textit{Mapper Lifting} (Graph to Hypergraph) by
Halley Fritze, Marissa Masden
\end{itemize}

\section{Conclusion}

This white paper presented the motivation and outcomes of the organization of the 2nd edition of the Topological Deep Learning Challenge hosted through the ICML 2024 Geometry-grounded Representation Learning and Generative Modeling (GRaM) Workshop. Challenge submissions implemented a wide variety of topological liftings between different pair of discrete (topological) domains, providing with a rich assortment of tools to infer and exploit higher-order structures. We hope that this community effort will help bridging the gap between TDL and most of the current datasets, stored as pointclouds or graphs. Therefore, the methods implemented in this challenge can potentially foster research and further methodological benchmarks in this growing TDL field.\footnote{In fact, all challenge submissions are by design compatible with \texttt{TopoBenchmarkX} framework~\cite{telyatnikov2024topobenchmarkx}, which easily allows the use of a diverse set of datasets, models and topological liftings.}

Last, but not least, we remark that the participation statistics of this 2nd challenge edition almost doubled the numbers of the previous ICML 2023 TDL Challenge~\cite{papillon2023icml} --both in terms of participants and submissions--, indicating a notable increase in interest in the TDL field. 

\clearpage
\onecolumn
\begin{longtable}[c]{L{4.2cm} L{0.7cm} L{4.2cm} C{1.2cm} C{1.2cm} C{1.2cm} C{1.2cm}}
%\begin{longtable}{l|l|l|l|l|l|l}
%\toprule
\textbf{Lifting name} & \textbf{ACs} & \textbf{Authors} & \textbf{Source domain} & \textbf{Dest. domain} & \textbf{Feat.-based} & \textbf{Conn.-based} \\
\bottomrule
Independent Sets Lifting & 1,4 & Frederic Wantiez & G & SC &  & \checkmark \\ \midrule
Neighborhood Lifting & 1,4 & Jonas Verhellen & G & CC &  & \checkmark \\
\midrule
Neighborhood/Dowker Lifting & 1,4 & Pavel Snopov & G & SC &  & \checkmark \\
\midrule
Vietoris-Rips Lifting & 1,3 & Jonas Verhellen & G & SC & \checkmark &  \\
\midrule
Graph Induced Lifting & 1,4 & Jonas Verhellen & G & SC &  & \checkmark \\
\midrule
Line Lifting & 1,4 & Pavel Snopov & G & SC &  & \checkmark \\
\midrule
Eccentricity Lifting & 1,4 & Jonas Verhellen & G & SC &  & \checkmark \\
\midrule
DnD Lifting & 1,3 & Jonas Verhellen & G & SC & \checkmark &  \\
\midrule
CellEncoding Lifting & 2,4 & Alexander Weers & CC & G &  & \checkmark \\
\midrule
KNN Graph Lifting & 2,3 & Frederic Wantiez & PC & SC & \checkmark &  \\
\midrule
Molecule Ring-Based Lifting & 1,4 & Bertran Miquel-Oliver, Manel Gil-Sorribes, Alexis Molina, Victor Guallar & G & CC &  & \checkmark \\
\midrule
Molecule Ring \& Close Atoms Lifting & 2,3 & Bertran Miquel-Oliver, Manel Gil-Sorribes, Alexis Molina, Victor Guallar & G & CCC & \checkmark & \checkmark \\
\midrule
Vietoris–Rips Lifting & 1,3 & Matouš Elphick & PC & SC & \checkmark &  \\
\midrule
Delaunay Lifting & 1,3 & Pavel Snopov & PC & SC & \checkmark &  \\
\midrule
KNN Lifting & 2,3 & Hongwei Jin & PC & G & \checkmark &  \\
\midrule
Witness Lifting & 1,3 & Pavel Snopov, German Magai & PC & SC & \checkmark &  \\
\midrule
Molecule Ring \& Functional Lifting & 2,3 & Bertran Miquel-Oliver, Manel Gil-Sorribes, Alexis Molina, Victor Guallar & G & CCC & \checkmark & \checkmark \\
\midrule
Alpha Complex Lifting & 1,3 & Theodore Long & PC & SC & \checkmark &  \\
\midrule
Expander Hypergraph Lifting & 2,4 & Julian Suk, Patryk Rygiel & G & HG &  & \checkmark \\
\midrule
Cy2C Lifting & 1,4 & Yun Young Choi, Minho Lee & G & SC &  & \checkmark \\
\midrule
N-Hop Lifting & 2,4 & Martin Carrasco & G & CCC &  & \checkmark \\
\midrule
Coface Lifting & 2,4 & Martin Carrasco & SC & CCC &  & \checkmark \\
\midrule
Kernel Lifting & 2,3 & Alexander Nikitin & G & HG & \checkmark & \checkmark \\
\midrule
Matroid Lifting & 2,4 & Giordan Escalona & G & CCC &  & \checkmark \\
\midrule
Forman-Ricci Curvature Coarse Geometry Lifting & 2,4 & Michael Banf, Dominik Filipiak, Max Schattauer, Liliya Imasheva & G & HG &  & \checkmark \\
\midrule
Voronoi Lifting & 2,3 & Julian Suk, Patryk Rygiel & PC & HG & \checkmark &  \\
\midrule
Feature-Based Rips Complex & 1,3 & Theodore Long & PC & SC & \checkmark &  \\
\midrule
Protein Close Residues Lifting & 2,3 & Bertran Miquel-Oliver, Manel Gil-Sorribes, Alexis Molina, Victor Guallar & G & HG & \checkmark & \checkmark \\
\midrule
Random Walks Lifting & 2,3 & Nikos Kanakaris, Veljko Kovač, Nesreen K Ahmed, Paul Bogdan, Andrei Irimia & PC & G & \checkmark &  \\
\midrule
Node Attribute Lifting & 2,3 & Yu Demi Qin, Graham Johnson & G & HG & \checkmark &  \\
\midrule
Neighbourhood Complex Lifting & 1,4 & Martin Carrasco & G & SC &  & \checkmark \\
\midrule
PointCloud to Graph Protein Lifting & 2,3 & Bertran Miquel-Oliver, Manel Gil-Sorribes, Alexis Molina, Victor Guallar & PC & G & \checkmark &  \\
\midrule
Path-based Lifting & 1,4 & Salvish Goomanee & G & CCC &  & \checkmark \\
\midrule
Directed Flag Complex Lifting & 2,3 & Thomas Gebhart & G & SC &  & \checkmark \\
\midrule
Mixture of Gaussians + MST Lifting & 2,3 & Sebastian Mežnar, Boshko Koloski, Blaž Škrlj & PC & HG & \checkmark &  \\
\midrule
Node centrality Lifting & 2,4 & Michael Banf, Dominik Filipiak, Max Schattauer, Liliya Imasheva & G & HG &  & \checkmark \\
\midrule
Universal Strict Lifting & 2,4 & Alvaro Martinez & HG & CCC &  & \checkmark \\
\midrule
Mapper Lifting & 2,4 & Halley Fritze, Marissa Masden & G & HG &  & \checkmark \\
\midrule
Modularity Maximization Lifting & 2,3 & Valentina Sánchez & G & HG & \checkmark & \checkmark \\
\midrule
Random Flag Complex Lifting & 1,3 & Martin Carrasco & PC & SC & \checkmark &  \\
\midrule
Path lifting & 2,4 & Pierrick Leroy, Marco Nurisso, Francesco Vaccarino & G & HG &  & \checkmark \\
\midrule
PointNet++ Lifting & 2,3 & Julian Suk, Patryk Rygiel & PC & HG & \checkmark &  \\
\midrule
Ball-Pivoting Lifting & 1,3 & Katrina Agate & PC & SC & \checkmark &  \\
\midrule
Spin Lifting & 1,4 & Pengfei Bai & G & PC &  & \checkmark \\
\midrule
Simplicial Paths Lifting & 2,4 & Manuel Lecha, Andrea Cavallo, Claudio Battiloro & G & CCC &  & \checkmark \\
\midrule
Hypergraph Heat Kernel Lifting & 1,4 & Matt Piekenbrock & HG & SC &  & \checkmark \\
\midrule
Tangential (Delaunay) Lifting & 1,3 & Maxim Beketov & PC & SC & \checkmark &  \\
\midrule
Probabilistic Clique Lifting & 2,4 & Alvaro Martinez & G & CCC &  & \checkmark \\
\midrule
Random Latent Clique Lifting & 1,4 & Mauricio Tec, Claudio Battiloro, George Dasoulas & G & SC &  & \checkmark \\
\midrule
Discrete Conf. Complex & 1,4 & Theodore Long & G & CC &  & \checkmark \\
\midrule
Mapper Lifting & 2,3 & Patrik Zajec & PC & G & \checkmark &  \\
\midrule
Spectral Lifting & 2,3 & Alessandro Salatiello & G & HG & \checkmark &  \\
\bottomrule
\caption[c]{List of valid submissions. Legend for domains: PC $\rightarrow$ pointcloud , G $\rightarrow$ graph, HG $\rightarrow$ hypergraph, \\ SC $\rightarrow$ simplicial, CC $\rightarrow$ cellular, and CCC $\rightarrow$ combinatorial.}
\label{tab:submissions}
\end{longtable}

\twocolumn

\bibliography{references}

\begin{thebibliography}{25}
\providecommand{\natexlab}[1]{#1}
\providecommand{\url}[1]{\texttt{#1}}
\expandafter\ifx\csname urlstyle\endcsname\relax
  \providecommand{\doi}[1]{doi: #1}\else
  \providecommand{\doi}{doi: \begingroup \urlstyle{rm}\Url}\fi

\bibitem[Barbarossa \& Sardellitti(2020)Barbarossa and Sardellitti]{barbarossa2020topological}
Barbarossa, S. and Sardellitti, S.
\newblock Topological signal processing over simplicial complexes.
\newblock \emph{IEEE Transactions on Signal Processing}, 68:\penalty0 2992--3007, 2020.

\bibitem[Battiloro et~al.(2023{\natexlab{a}})Battiloro, Sardellitti, Barbarossa, and Di~Lorenzo]{battiloro2023topological}
Battiloro, C., Sardellitti, S., Barbarossa, S., and Di~Lorenzo, P.
\newblock Topological signal processing over weighted simplicial complexes.
\newblock In \emph{ICASSP 2023-2023 IEEE International Conference on Acoustics, Speech and Signal Processing (ICASSP)}, pp.\  1--5. IEEE, 2023{\natexlab{a}}.

\bibitem[Battiloro et~al.(2023{\natexlab{b}})Battiloro, Spinelli, Telyatnikov, Bronstein, Scardapane, and Di~Lorenzo]{battiloro2023latent}
Battiloro, C., Spinelli, I., Telyatnikov, L., Bronstein, M., Scardapane, S., and Di~Lorenzo, P.
\newblock From latent graph to latent topology inference: differentiable cell complex module.
\newblock \emph{arXiv preprint arXiv:2305.16174}, 2023{\natexlab{b}}.

\bibitem[Battiston et~al.(2021)Battiston, Amico, Barrat, Bianconi, Ferraz~de Arruda, Franceschiello, Iacopini, K{\'e}fi, Latora, Moreno, et~al.]{battiston2021physics}
Battiston, F., Amico, E., Barrat, A., Bianconi, G., Ferraz~de Arruda, G., Franceschiello, B., Iacopini, I., K{\'e}fi, S., Latora, V., Moreno, Y., et~al.
\newblock The physics of higher-order interactions in complex systems.
\newblock \emph{Nature Physics}, 17\penalty0 (10):\penalty0 1093--1098, 2021.

\bibitem[Bern{\'a}rdez et~al.(2023)Bern{\'a}rdez, Telyatnikov, Alarc{\'o}n, Cabellos-Aparicio, Barlet-Ros, and Li{\`o}]{bernardez2023topological}
Bern{\'a}rdez, G., Telyatnikov, L., Alarc{\'o}n, E., Cabellos-Aparicio, A., Barlet-Ros, P., and Li{\`o}, P.
\newblock Topological network traffic compression.
\newblock In \emph{Proceedings of the 2nd Graph Neural Networking Workshop 2023}, pp.\  7--12, 2023.

\bibitem[Bick et~al.(2023)Bick, Gross, Harrington, and Schaub]{bick2023higher}
Bick, C., Gross, E., Harrington, H.~A., and Schaub, M.~T.
\newblock What are higher-order networks?
\newblock \emph{SIAM Review}, 65\penalty0 (3):\penalty0 686--731, 2023.

\bibitem[Bodnar(2023)]{bodnar2023topological}
Bodnar, C.
\newblock \emph{Topological deep learning: graphs, complexes, sheaves}.
\newblock PhD thesis, School of Technology, Faculty of Computer Science and Technology, University of Cambridge, 2023.

\bibitem[Bodnar et~al.(2021{\natexlab{a}})Bodnar, Frasca, Otter, Wang, Lio, Montufar, and Bronstein]{bodnar2021weisfeiler}
Bodnar, C., Frasca, F., Otter, N., Wang, Y., Lio, P., Montufar, G.~F., and Bronstein, M.
\newblock Weisfeiler and lehman go cellular: Cw networks.
\newblock \emph{Advances in neural information processing systems}, 34:\penalty0 2625--2640, 2021{\natexlab{a}}.

\bibitem[Bodnar et~al.(2021{\natexlab{b}})Bodnar, Frasca, Otter, Wang, Lio, Montufar, and Bronstein]{bodnar2021weisfeilercellular}
Bodnar, C., Frasca, F., Otter, N., Wang, Y., Lio, P., Montufar, G.~F., and Bronstein, M.
\newblock Weisfeiler and lehman go cellular: Cw networks.
\newblock \emph{Advances in neural information processing systems}, 34:\penalty0 2625--2640, 2021{\natexlab{b}}.

\bibitem[Bodnar et~al.(2021{\natexlab{c}})Bodnar, Frasca, Wang, Otter, Montufar, Lio, and Bronstein]{bodnar2021simplicial}
Bodnar, C., Frasca, F., Wang, Y., Otter, N., Montufar, G.~F., Lio, P., and Bronstein, M.
\newblock Weisfeiler and lehman go topological: Message passing simplicial networks.
\newblock In \emph{International Conference on Machine Learning}, pp.\  1026--1037. PMLR, 2021{\natexlab{c}}.

\bibitem[Elshakhs et~al.(2024)Elshakhs, Deliparaschos, Charalambous, Oliva, and Zolotas]{elshakhs2024comprehensive}
Elshakhs, Y.~S., Deliparaschos, K.~M., Charalambous, T., Oliva, G., and Zolotas, A.
\newblock A comprehensive survey on delaunay triangulation: Applications, algorithms, and implementations over cpus, gpus, and fpgas.
\newblock \emph{IEEE Access}, 2024.

\bibitem[Hagberg \& Conway(2020)Hagberg and Conway]{hagberg2020networkx}
Hagberg, A. and Conway, D.
\newblock Networkx: Network analysis with python.
\newblock \emph{URL: https://networkx. github. io}, 2020.

\bibitem[Hajij \& Istvan(2021)Hajij and Istvan]{hajij2021topological}
Hajij, M. and Istvan, K.
\newblock Topological deep learning: Classification neural networks.
\newblock \emph{arXiv preprint arXiv:2102.08354}, 2021.

\bibitem[Hajij et~al.(2022{\natexlab{a}})Hajij, Zamzmi, Papamarkou, Miolane, Guzm{\'a}n-S{\'a}enz, and Ramamurthy]{hajij2022higher}
Hajij, M., Zamzmi, G., Papamarkou, T., Miolane, N., Guzm{\'a}n-S{\'a}enz, A., and Ramamurthy, K.~N.
\newblock Higher-order attention networks.
\newblock \emph{arXiv preprint arXiv:2206.00606}, 2\penalty0 (3):\penalty0 4, 2022{\natexlab{a}}.

\bibitem[Hajij et~al.(2022{\natexlab{b}})Hajij, Zamzmi, Papamarkou, Miolane, Guzm{\'a}n-S{\'a}enz, Ramamurthy, Birdal, Dey, Mukherjee, Samaga, et~al.]{hajij2022topologicalDL}
Hajij, M., Zamzmi, G., Papamarkou, T., Miolane, N., Guzm{\'a}n-S{\'a}enz, A., Ramamurthy, K.~N., Birdal, T., Dey, T.~K., Mukherjee, S., Samaga, S.~N., et~al.
\newblock Topological deep learning: Going beyond graph data.
\newblock \emph{arXiv preprint arXiv:2206.00606}, 2022{\natexlab{b}}.

\bibitem[Hajij et~al.(2023)Hajij, Zamzmi, Papamarkou, Guzman-Saenz, Birdal, and Schaub]{hajij2023combinatorial}
Hajij, M., Zamzmi, G., Papamarkou, T., Guzman-Saenz, A., Birdal, T., and Schaub, M.~T.
\newblock Combinatorial complexes: bridging the gap between cell complexes and hypergraphs.
\newblock In \emph{2023 57th Asilomar Conference on Signals, Systems, and Computers}, pp.\  799--803. IEEE, 2023.

\bibitem[Hajij et~al.(2024)Hajij, Papillon, Frantzen, Agerberg, AlJabea, Ballester, Battiloro, Bern{\'a}rdez, Birdal, Brent, et~al.]{hajij2024topox}
Hajij, M., Papillon, M., Frantzen, F., Agerberg, J., AlJabea, I., Ballester, R., Battiloro, C., Bern{\'a}rdez, G., Birdal, T., Brent, A., et~al.
\newblock Topox: a suite of python packages for machine learning on topological domains.
\newblock \emph{arXiv preprint arXiv:2402.02441}, 2024.

\bibitem[Hoppe \& Schaub(2024)Hoppe and Schaub]{hoppe2024representing}
Hoppe, J. and Schaub, M.~T.
\newblock Representing edge flows on graphs via sparse cell complexes.
\newblock In \emph{Learning on Graphs Conference}, pp.\  1--1. PMLR, 2024.

\bibitem[Papamarkou et~al.(2024)Papamarkou, Birdal, Bronstein, Carlsson, Curry, Gao, Hajij, Kwitt, Lio, Di~Lorenzo, et~al.]{papamarkou2024position}
Papamarkou, T., Birdal, T., Bronstein, M.~M., Carlsson, G.~E., Curry, J., Gao, Y., Hajij, M., Kwitt, R., Lio, P., Di~Lorenzo, P., et~al.
\newblock Position: Topological deep learning is the new frontier for relational learning.
\newblock In \emph{Forty-first International Conference on Machine Learning}, 2024.

\bibitem[Papillon et~al.(2023{\natexlab{a}})Papillon, Hajij, Myers, Frantzen, Zamzmi, Jenne, Mathe, Hoppe, Schaub, Papamarkou, et~al.]{papillon2023icml}
Papillon, M., Hajij, M., Myers, A., Frantzen, F., Zamzmi, G., Jenne, H., Mathe, J., Hoppe, J., Schaub, M., Papamarkou, T., et~al.
\newblock Icml 2023 topological deep learning challenge: design and results.
\newblock In \emph{Topological, Algebraic and Geometric Learning Workshops 2023}, pp.\  3--8. PMLR, 2023{\natexlab{a}}.

\bibitem[Papillon et~al.(2023{\natexlab{b}})Papillon, Sanborn, Hajij, and Miolane]{papillon2023architectures}
Papillon, M., Sanborn, S., Hajij, M., and Miolane, N.
\newblock Architectures of topological deep learning: A survey of message-passing topological neural networks.
\newblock \emph{arXiv preprint arXiv:2304.10031}, 2023{\natexlab{b}}.

\bibitem[Scarselli et~al.(2009)Scarselli, Gori, Tsoi, Hagenbuchner, and Monfardini]{scarselli2009}
Scarselli, F., Gori, M., Tsoi, A.~C., Hagenbuchner, M., and Monfardini, G.
\newblock The graph neural network model.
\newblock \emph{IEEE Transactions on Neural Networks}, 20\penalty0 (1):\penalty0 61--80, 2009.
\newblock \doi{10.1109/TNN.2008.2005605}.

\bibitem[Schaub et~al.(2020)Schaub, Benson, Horn, Lippner, and Jadbabaie]{schaub2020random}
Schaub, M.~T., Benson, A.~R., Horn, P., Lippner, G., and Jadbabaie, A.
\newblock Random walks on simplicial complexes and the normalized hodge 1-laplacian.
\newblock \emph{SIAM Review}, 62\penalty0 (2):\penalty0 353--391, 2020.

\bibitem[Telyatnikov et~al.(2024)Telyatnikov, Bernardez, Montagna, Vasylenko, Zamzmi, Hajij, Schaub, Miolane, Scardapane, and Papamarkou]{telyatnikov2024topobenchmarkx}
Telyatnikov, L., Bernardez, G., Montagna, M., Vasylenko, P., Zamzmi, G., Hajij, M., Schaub, M.~T., Miolane, N., Scardapane, S., and Papamarkou, T.
\newblock Topobenchmarkx: A framework for benchmarking topological deep learning.
\newblock \emph{arXiv preprint arXiv:2406.06642}, 2024.

\bibitem[Xu et~al.(2022)Xu, Li, Ni, Zhang, and Chen]{xu2022groupnet}
Xu, C., Li, M., Ni, Z., Zhang, Y., and Chen, S.
\newblock Groupnet: Multiscale hypergraph neural networks for trajectory prediction with relational reasoning.
\newblock In \emph{Proceedings of the IEEE/CVF Conference on Computer Vision and Pattern Recognition}, pp.\  6498--6507, 2022.

\end{thebibliography}
\bibliographystyle{icml2024}
\end{document}